\documentclass[conference]{IEEEtran}
\IEEEoverridecommandlockouts

\usepackage{color}
\usepackage{xspace}
\usepackage{url}
\usepackage[font={footnotesize},labelfont=bf]{caption,subfig} 
\usepackage{pifont} 
\usepackage{amsmath}

\usepackage{silence} 
\usepackage[normalem]{ulem} 

\usepackage{enumitem} 
\setlist{nosep} 

\usepackage{wrapfig}
\usepackage{rotating} 

\usepackage{booktabs} 

\usepackage{array} 
\newcommand{\PreserveBackslash}[1]{\let\temp=\\#1\let\\=\temp}
\newcolumntype{C}[1]{>{\PreserveBackslash\centering}p{#1}}
\newcolumntype{R}[1]{>{\PreserveBackslash\raggedleft}p{#1}}
\newcolumntype{L}[1]{>{\PreserveBackslash\raggedright}p{#1}}

\usepackage{titlesec}
\titlespacing\section{0pt}{4pt plus 4pt minus 2pt}{2pt plus 2pt minus 2pt}
\titlespacing\subsection{0pt}{4pt plus 4pt minus 2pt}{2pt plus 2pt minus 2pt}
\titlespacing\subsubsection{0pt}{4pt plus 4pt minus 2pt}{2pt plus 2pt minus 2pt}

\newcommand{\sys}[0]{Ares\xspace}

\newcommand{\cifar}[0]{CIFAR-10\xspace}

\newcommand{\vggs}[0]{VGG11\xspace}

\newcommand{\resnets}[0]{ResNet-18\xspace}
\newcommand{\resnetl}[0]{ResNet-50\xspace}

\newif\ifsubmit
\submitfalse

\ifsubmit
    \newcommand{\amir}[1]{}
    \newcommand{\kevin}[1]{}
    \newcommand{\pratik}[1]{}
    \newcommand{\farhan}[1]{}
    
    \newcommand{\todo}[1]{}
\else
    
    \newcommand{\amir}[1]{\textcolor{blue}{Amir: #1}}
    \newcommand{\kevin}[1]{\textcolor{green}{Kevin: #1}}
    \newcommand{\pratik}[1]{\textcolor{magenta}{Pratik: #1}}
    \newcommand{\farhan}[1]{\textcolor{orange}{Farhan: #1}}
    
    \newcommand{\todo}[1]{\textcolor{red}{TODO: #1}}
\fi

\newcommand{\paratitle}[1]{\noindent\textbf{\textit{#1.}}\xspace}

\newcommand{\ie}[0]{\emph{i.e.,}\xspace}
\newcommand{\etal}[0]{\emph{et~al.}\xspace}
\newcommand{\eg}[0]{\emph{e.g.,}\xspace}

\usepackage{cleveref} 

\crefformat{section}{\S#2#1#3} 
\crefformat{subsection}{\S#2#1#3}
\crefformat{subsubsection}{\S#2#1#3}

\usepackage{cite}
\usepackage{amsmath,amssymb,amsfonts}
\usepackage{algorithmic}
\usepackage{graphicx}
\usepackage{textcomp}
\usepackage{xcolor}
\usepackage{lipsum}
\usepackage{array,multirow}

\def\BibTeX{{\rm B\kern-.05em{\sc i\kern-.025em b}\kern-.08em
    T\kern-.1667em\lower.7ex\hbox{E}\kern-.125emX}}
\begin{document}

\title{\sys: A System-Oriented Wargame Framework for Adversarial ML
}

\author{
\IEEEauthorblockN{Farhan Ahmed}
\IEEEauthorblockA{\textit{Stony Brook University} \\
farhaahmed@cs.stonybrook.edu}
\and
\IEEEauthorblockN{Pratik Vaishnavi}
\IEEEauthorblockA{\textit{Stony Brook University} \\
pvaishnavi@cs.stonybrook.edu}
\and
\IEEEauthorblockN{Kevin Eykholt}
\IEEEauthorblockA{\textit{IBM Research} \\
kheykholt@ibm.com}
\and
\IEEEauthorblockN{Amir Rahmati}
\IEEEauthorblockA{\textit{Stony Brook University} \\
amir@cs.stonybrook.edu}
}

\maketitle

\begin{abstract}
Since the discovery of adversarial attacks against machine learning models nearly a decade ago, research on adversarial machine learning has rapidly evolved into an eternal war between defenders, who seek to increase the robustness of ML models against adversarial attacks, and adversaries, who seek to develop better attacks capable of weakening or defeating these defenses. This domain, however, has found little buy-in from ML practitioners, who are neither overtly concerned about these attacks affecting their systems in the real world nor are willing to trade off the accuracy of their models in pursuit of robustness against these attacks.

In this paper, we motivate the design and implementation of \sys, an evaluation framework for adversarial ML that allows researchers to explore attacks and defenses in a realistic wargame-like environment. \sys frames the conflict between the attacker and defender as two agents in a reinforcement learning environment with opposing objectives. This allows the introduction of system-level evaluation metrics such as time to failure and evaluation of complex strategies such as moving target defenses. We provide the results of our initial exploration involving a white-box attacker against an adversarially trained defender.
\end{abstract}


\section{Introduction}
The mass adoption of AI-powered systems has motivated re-examination of the reliability, privacy, and security of AI algorithms. With respect to security, it was discovered early on that image based AI algorithms are vulnerable to a class of \textit{adversarial evasion} attacks~\cite{szegedy2014intriguing, goodfellow2014explaining}. In such attacks, an adversary introduces a small amount of noise, imperceptible to the human eye, in order to reliably induce misclassification errors during inference. Since its discovery, a large body of research has proposed numerous empirical defense strategies such as transforming the model's inputs~\cite{guo2017countering}, modifying the neural network architecture~\cite{gu2015deep}, and training the network on an alternative training dataset~\cite{chen2018improving}. Despite the vast number of works, both in developing new adversarial attacks and proposing new defenses, including robust physical world attacks~\cite{eykholt2018robust}, the adversarial threat model remains unmotivating to ML practitioners. In a small industry survey, Kumar~\etal~\cite{9283867}, discovered that while most organizations surveyed were aware of adversarial examples, they remarked ``\textit{This [adversarial ML] looks futuristic}'' and lack tools in place to study and mitigate such attacks.

We argue that two key issues hinder the acceptance of adversarial evasion attacks as a threat: (1) the unmotivating threat model used by most prior work and (2) the lack of tools for evaluating complex adversarial attacker and defender interactions. Following Kerckhoffs's principle, adversarial attacks and defenses have mainly been studied using a white-box threat model, \ie full knowledge of the network and defense parameters. Under this lens, many proposed defenses were shown to be ineffective as an attacker with perfect knowledge could adapt to the defense~\cite{Athalye2018ObfuscatedGG}. However, such a strong threat model can only be replicated by attackers with insider access to the AI algorithm and training data. In real deployment scenarios, an organization is primarily concerned about the security of its AI systems against outside attackers. 

Despite the lack of recognition of adversarial ML as a threat, there has been a rise in adversarial attack libraries that enable ML practitioners to study the current state-of-the-art attack and defense algorithms. Some examples include University of Toronto's \textit{CleverHans}~\cite{papernot2018cleverhans}, MIT's \textit{robustness package}~\cite{robustness}, University of Tübingen's \textit{Foolbox}~\cite{rauber2017foolboxnative}, and IBM's \textit{Adversarial Robustness Toolbox (ART)}~\cite{art2018}. Each library defines a unified framework through which practitioners can evaluate the effectiveness of an attack or defense using their own AI systems. Unfortunately, such evaluations are limited by nature as the evaluated threat model is limited by the attack algorithm. Furthermore, both the attacker and defender are assumed static. They do not modify their behavior based on the actions of the other and, as such, the reported effectiveness is misleading and does not translate into a meaningful notion of effectiveness in the real world.

In this paper, we describe a new evaluation framework, \sys, which represents adversarial attack scenarios as a complex, dynamic interaction between the attacker and defender. We explore the conflict between the attacker and defender as two independent agents in a reinforcement learning (RL) environment with opposing objectives, creating a richer and more realistic environment for adversarial ML evaluation. By utilizing this RL-environment, we are able to tweak the attacker or defender's strategy (the RL policy) to be static, randomized, or even learnable. \sys also allows the investigation of both white-box and black-box threat models, drawing inspiration from the limitations of prior evaluations. 

For its debut, we have used \sys to re-examine the security of the ensemble/moving target defense (MTD) framework in a white-box scenario and highlight the vulnerability of this setup. Using different combinations of naturally trained and adversarially trained models, an \sys evaluation finds that, in general, the attacker always wins and adversarial training can only slightly delay the attacker's success. As prior work discusses, the attacker's success is largely due to the transferability of adversarial examples~\cite{goodfellow2014explaining}. We investigate this phenomenon more thoroughly through the lens of \sys and discover that the shared loss gradients between networks, regardless of training method or model architecture, is the main culprit. We then discuss how MTDs could be improved based on this discovery and our next steps towards evaluating MTDs and other prior works in a black-box threat model through \sys.

In this paper we make the following contributions:
\begin{itemize}
    \item We develop \sys, an RL-based evaluation framework for adversarial ML that allows researchers to explore attack/defense strategies at a system level.
    \item Using \sys, we re-examine ensemble/moving target defense strategies under the white-box threat model and show that the root cause of this failure is due to the shared loss gradient between the networks.
\end{itemize}

The \sys framework is publicly available at \texttt{\url{https://github.com/Ethos-lab/ares}} as we continue development for additional features and improvement.

\section{Background \& Related Work}

\paratitle{Adversarial Evasion Attacks}
Prior works have uncovered several classes of vulnerabilities for ML models and designed attacks to exploit them~\cite{chakraborty2018adversarial}. In this paper, we focus on one such class of attacks known as \textit{evasion attacks}. In an evasion attack, the adversary's goal is to generate an ``adversarial example'' -- a carefully perturbed input that causes misclassification. Evasion attacks against ML models have been developed to suit a wide range of scenarios. \textit{White-box attacks}~\cite{szegedy2014intriguing,goodfellow2014explaining,kurakin2016adversarial,croce2020reliable} assume full knowledge of/access to the model, including but not limited to model's architecture, parameters, gradients, and training data. Such attacks, although extremely potent, are mostly impractical in real-world scenarios~\cite{pierazzi2020intriguing} as the ML models used in commercial systems are usually hidden underneath a layer of system/network security measures. Focusing on strengthening these security measures not only provides improved protection for the underlying ML models against white-box attacks, it also improves the overall security posture of the system, and hence, is often a more practical and desirable approach.
\textit{Black-box attacks}~\cite{papernot2017practical,eykholt2018robust,brendel2018decision,shi2019curls,dong2019evading,chen2020hopskipjumpattack,croce2020minimally}, on the other hand, only assume query access to the target ML models. Such a threat model offers a more practical assumption as several consumer facing ML models provide this access to their users~\cite{azurecv,googlecv,awsrekognition,clarifai}. 


\paratitle{Defenses against Evasion Attacks}
A wide range of strategies to address the threat of adversarial evasion attacks have also been proposed. One line of works look at tackling this issue at test-time~\cite{gu2015deep,guo2017countering,meng2017magnet,chen2018improving,liao2018defense}. These works usually involve variations of a preprocessing step that filters out the adversarial noise from the input before feeding it to the ML model. These defenses, however, have been shown to convey a false sense of security and so, been easily broken using adaptive attacks~\cite{Athalye2018ObfuscatedGG}. 

Another popular strategy involves re-training the model using a robustness objective~\cite{madry2018towards,zhang2019theoretically,xie2019feature}. The defenses that employ this strategy show promise as they have (so far) stood strong in the face of adaptive adversaries. All the defenses discussed so far belong in the broad category of \textit{empirical defenses}. These defenses only provide empirical guarantees of robustness and may not be secure against a future attack. Another line of works look at developing methods that can train certifiably robust ML models~\cite{cohen2019certified,salman2019provably,zhai2020macer}. These models can offer formal robustness guarantees against any attacker with a pre-defined budget.

\paratitle{Defenses based on Ensembling}
One commonly known property of adversarial examples is that they can similarly fool models independently trained on the same data~\cite{goodfellow2014explaining}. Adversaries can exploit this property by training a surrogate model to generate adversarial examples against a target model. This, in fact, is a popular strategy used by several black-box attacks~\cite{papernot2017practical,shi2019curls,dong2019evading}. Tram{\`e}r~\etal~\cite{tramer2018ensembletraining} use this property to improve the black-box robustness of models trained using the single-step attack version of adversarial training. At each training iteration, source of adversarial examples is randomly selected from an ensemble containing the currently trained model and a set of pre-trained models. Other works~\cite{abbasi2017robustness,verma2019error,pang2019improving,sen2019empir} propose strategies for training a diverse pool of models so that it is difficult for an adversarial example to transfer across the majority of them. Aggregating the outputs of these diverse models should therefore yield improved robustness. This ensemble diversity strategy, however, has been shown to be ineffective~\cite{he2017adversarial,tramer2020adaptive}. 
In similar vain, some prior works~\cite{sengupta2018mtdeep,roy2019moving} propose use of  ensemble of models as a moving target defense where, depending on the MTD strategy, the attacker may face a different target model in each encounter.
These works, unfortunately, suffer from the same shortcomings of the ensemble methods.

\paratitle{Adversarial ML Libraries} To facilitate research into machine learning security, multiple research groups and organizations have developed libraries to assist in development and evaluation of adversarial attacks and defenses. Most notably of these works are University of Toronto's \textit{CleverHans}~\cite{papernot2018cleverhans}, MIT's \textit{robustness package}~\cite{robustness}, University of Tübingen's \textit{Foolbox}~\cite{rauber2017foolboxnative}, and IBM's \textit{Adversarial Robustness Toolbox (ART)}~\cite{art2018}.

These efforts are orthogonal to our framework. While \sys focuses on evaluating various attacker and defender strategies against one another across multiple scenarios, these libraries focus primarily on facilitating implementation of new attacks and defenses and benchmarking them against existing ones. In this paper, we use the Projected Gradient Descent (PGD) attack from IBM's ART library as our main adversarial evaluation criteria.

\section{\sys Framework}

\begin{figure*}[t]
\centering
\caption{An overview figure of the \sys framework. The RL-environment consists of three components: the evaluation scenario, attacker agent, and defender agent. Both the attacker and defender obtain information from the evaluation scenario and battle against each other with their own predefined objectives based on the end states.}
\includegraphics[width=0.75\textwidth]{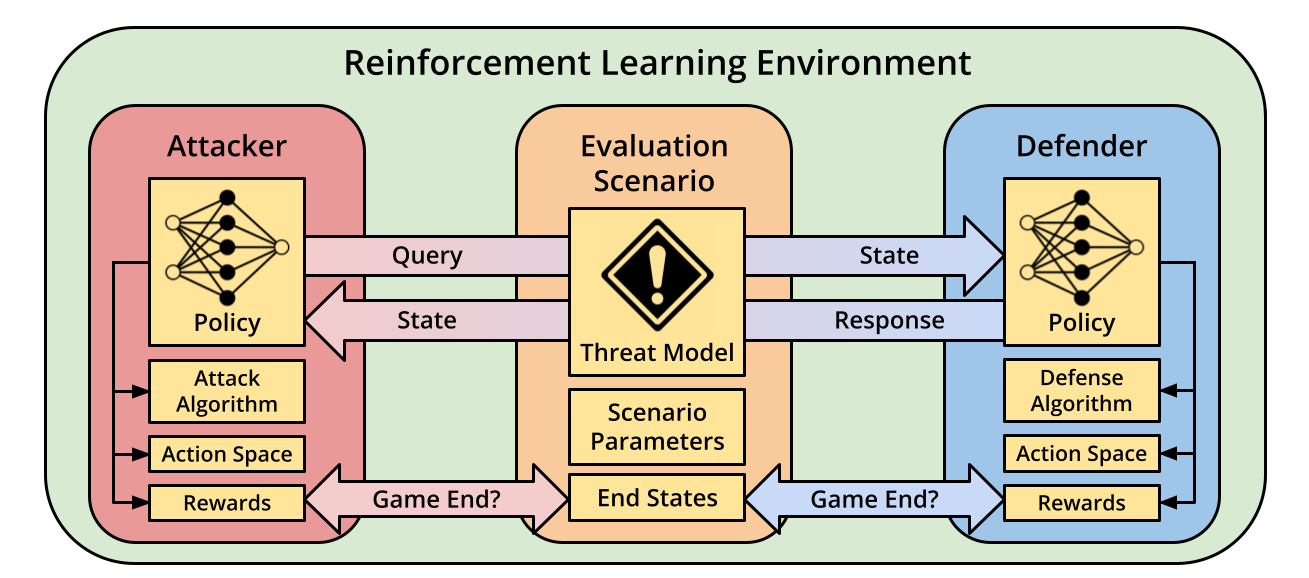}
\label{fig:framework}
\vspace{-1.75em}
\end{figure*}

In this section, we provide an overview of the \sys framework. As seen in Figure \ref{fig:framework}, \sys adapts the adversarial attack/defense problem into an RL-environment consisting of three main components: (1) the evaluation scenario, (2) the attacker agent, and (3) the defender agent. Once each component has been defined by the user, \sys executes a series of competitions between the attacker and defender. Each competition is modeled as a turn-based game with the attacker taking the first move and defender responding during each round. To do this, \sys freezes one agent's policy, either the attacker or defender, and treats it as part of the RL-environment, while the opposing agent makes their move. The current active agent (\ie attacker or defender) interacts with the RL-environment (specifically the evaluation scenario) and takes actions based on the current game state. This setup enables us to tune and evaluate each agent individually. The game will end when the active agent achieves its end condition (\eg the attacker has successfully fooled the model, the defender has detected the attacker, or the defender has resisted the attack for a set number of rounds). The final output of the system is a series of metrics that reflect the effectiveness of the attacker and defender for the current scenario.

\subsection{Evaluation Scenario}
In \sys, the evaluation scenario is defined by the threat model, scenario parameters, and end states. Currently, \sys supports the following threat models:
\begin{enumerate}
    \item \textit{White-box}: In this threat model, the attacker is provided the complete loss gradient when issuing queries to the defender. If a defense uses gradient obfuscation, the attacker is provided approximated gradients based on the adaptive measures from Athalye~\etal~\cite{Athalye2018ObfuscatedGG}.
    \item \textit{Soft Black-box}: In this threat model, the attacker is provided only the probabilistic outputs of the model.
    \item \textit{Hard Black-box}: In this threat model, the attacker is provided only the model's label prediction.
\end{enumerate}
We've used these threat models in our current implementation as they represent the most common ones found in the literature. The threat model is used to determine the amount of information provided to the attacker by the defender.

The scenario parameters consist of the shared evaluation parameters used by both the attacker and defender. This includes the dataset, number of competitions executed, and the time limit (\ie number of steps). The end states define the ending conditions for a game round. In addition to ending the game when time expires, the environment will also end the game if the active component (\ie attacker or defender) has achieved its win/loss condition. For example, the game may end if the attacker has successfully generated an adversarial sample that the defender misclassifies.

\subsection{Attacker Agent}
To define an attacker, the user must define the policy, attack algorithm, action space, and reward function. The policy utilizes the other three subcomponents and defines the attacker's behavior and overall strategy. The policy can be static, randomized, or even learnable and dynamically changing throughout the game. Often, the definition of the attack algorithm will depend on the specific threat model as current adversarial attacks assume that the loss gradient is either present or must be approximated based on model outputs. We note that the attacker is not limited to a single attack algorithm.

The action space defines the attacker's potential moves. In scenarios against a static defense (\ie the defender does not take measures to monitor or adapt to the attack actions), often the action space will simply involve issuing a query to the defender and using its response to take an attack optimization step. However, when the defender attempts to adapt to or detect malicious actions, the attacker's action space may include evasive actions such as issuing queries that do not advance the attack in any way. In this case, the attacker will learn its policy through reinforcement learning to determine its action based on the current state obtained from the environment. 

Finally, the reward function defines a measure to evaluate the effectiveness of the attacker. Most prior work measure the effectiveness of an attack by the number of successfully attacked samples in the dataset, usually denoted as the adversarial accuracy or robustness. As such, adversarial accuracy is one of the two default measures of our framework. As a second measure of effectiveness, we also measure the number of attack steps required to perform a successful attack. In certain scenarios such as an attack with limited queries, it may be sufficient for a defense to simply hinder the attack rather than completely prevent it. The reward function can also include additional bonuses or penalties in support of scenarios where the defender can respond. For example, if the defender has detection capabilities, the attacker could be penalized if detected. We choose to define the reward function within the attacker component rather than the environment to allow for asymmetrical reward definitions for the attacker and defender. It also allows for the definition of multiple attacker agents.

\subsection{Defender Agent}
To define a defender, the user must similarly define the policy, action space, defense algorithm, and reward function.
Similar to the attacker agent, the policy utilizes the other three subcomponents to define the defender's strategy throughout the game and can be static, randomized, or learnable.
The defense algorithm opposes the attacker's attack algorithm and similarly depends on the threat model. Just like the attacker, the defender is not limited to a single defense algorithm.

The action space of the defender dictates how the defender responds to a query. As a basic requirement, the defender must always respond truthfully and cannot lie such that the output prediction is invalid. This requirement is motivated by the fact that in a real scenario, the defense is unaware of the query source, and thus should respond as if the source is benign. Furthermore, if the defender was able to detect the attacker, we could end the game prematurely and penalize the attacker, rather than continue the game with deceptive responses from the defender. Based on prior works, the action space can include preprocessing of the input or postprocessing of the output. Of interest are scenarios with a non-static defender as the action space could include steps to detect if the current query is a query used for an attack optimization. Just like the attacker, in the case of a non-static defense, the defender will learn a policy through reinforcement learning to determine its actions based on the current state obtained from the environment.

The reward function defines a measure to evaluate the effectiveness of the defender. The defender's reward function will often use the same metrics as the attacker. Whereas by default, the attacker wants to minimize the adversarial accuracy and time required to succeed, the defender wants to maximize the values. Similarly, if the defense is not static, a reward can be given whenever the defense successfully responds to an attack optimization query. As before, the reward is defined specific to the defender to allow for asymmetrical reward definitions to also allow for multiple defenders.

\subsection{Framework Implementation}

To implement \sys, we utilize Open AI's \textit{Gym}~\cite{gym} library to create the RL-environment composed of the evaluation scenario, attacker agent, and defender agent. Through integration with the Adversarial Robustness Toolbox (\textit{ART}~\cite{art2018}), our framework supports numerous adversarial attacks and defenses by default. In addition, users can implement their own attack or defenses algorithms rather than rely on ART's implementations if customization is required. For example, in this paper, we implemented the moving target defense (MTD) using custom code. Finally, in the current implementation, our framework evaluates scenarios with a single attacker and single defender agent. We are currently working to enable support for multiple attacker and defender agents.

\section{Results}
To showcase the capabilities of \sys, we look back at the use of ensemble models as an adversarial defense. Prior work~\cite{goodfellow2014explaining} in adversarial evasion attacks found that adversarial samples were \textit{transferable}. That is to say, an adversarial sample for one model was highly likely to also be adversarial for another model regardless of architecture. Therefore, ensembling of models was assumed to do little towards create adversarially robust classifiers. In this section, we use \sys to evaluate the effectiveness of the moving target defense (MTD) architecture under white-box threat model. Specifically, we study this defense using both naturally trained and adversarially trained models of both the same and different architectures. We discuss the evaluation results obtained from \sys and demonstrate how these results reveal the underlying reason for adversarial transferability.

\subsection{Experimental Setup}

For the environment parameters, we set the threat model to a white-box threat model, We use the \cifar dataset and define the end states to be either when the attacker causes misclassification on a sample originally correctly classified by the all of the defender's models or when time expires. Each game is run for a maximum of 20 steps.

For the attack algorithm, we used the Projected Gradient Descent (PGD) attack~\cite{madry2018towards} implementation from ART~\cite{art2018}. We set the attack budget to $\epsilon = 8/255$ and step size $\alpha = 2/255$, and perform an $\ell_{\infty}$-norm PGD attack. In each step, the attack performs a 1-step PGD attack using the state returned by the environment. As this is a white-box threat model, the environment state is the loss gradient of the defender's model with respect to the attacker's query. The action space of the attacker only contains the query action given that the defender is static. Finally, the attacker's reward is measured by the number of steps required to generate a misclassified sample.

For the defense algorithm, we use an MTD architecture consisting of some combination of up to six models. Using PyTorch~\cite{pytorch} implementations of the \resnets, \resnetl, and \vggs architectures, we train each architecture using both natural and adversarial training~\cite{madry2018towards}\footnote{For more information on the natural and adversarial accuracy for each model, refer to Table~\ref{tab:train-acc}.}. For adversarial training, we set $\epsilon = 8/255$, step size $\alpha = 2/255$, and perform a 10-step $\ell_{\infty}$-norm  PGD attack.  The defender uses a static policy that randomly selects one of the models in the MTD architecture and uses it to respond to the query. Finally, the defender's reward is measured by the number of steps elapsed before the attack succeeds.

\begin{table}[h!]
\centering
\caption{The natural and adversarial accuracy for all the pre-trained models used in our experiments. The experiments were conducted using the \cifar dataset. Each model was trained either naturally (N) or adversarially (A). For adversarial training and evaluation, we use an $\ell_{\infty}$-norm 10-step PGD with attack budget $\epsilon = 8/255$ and step size $\alpha = 2/255$.}
\resizebox{0.9\columnwidth}{!}{
\begin{tabular}{@{}lccc@{}}
\toprule
\multirow{2}{*}[-0.4em]{\textbf{Model}} & \multirow{2}{*}[-0.4em]{\textbf{Train Method}} & \multicolumn{2}{c}{\textbf{Accuracy}} \\
\cmidrule(l){3-4} 
 &  & \textbf{Natural} & \textbf{Adversarial} \\ \midrule
\multirow{2}{*}{\resnets} & N & 93.07\% & 2.25\% \\
 & A & 82.60\% & 51.59\% \\
\midrule
\multirow{2}{*}{\resnetl} & N & 93.65\% & 1.46\% \\
 & A & 83.73\% & 51.06\% \\
\midrule
\multirow{2}{*}{\vggs} & N & 92.39\% & 2.55\% \\
 & A & 76.94\% & 44.15\% \\
\bottomrule
\end{tabular}
}
\label{tab:train-acc}
\end{table}

\subsection{MTD with Natural and Adversarial Training}

\begin{table}[]
\centering
\caption{The number of rounds until the attacker succeeds in triggering misclassification by the defender using the \sys framework. For each of the three models (\resnets, \resnetl, and \vggs) we run the game with the MTD consisting of either only the naturally trained (N), only the adversarially trained (A), or a mixture of both (N + A). We perform 100 trials for each experiment and report the mean number of rounds with its 95\% confidence interval. The presence of a naturally trained model only weakens the robustness of the MTD while the network architecture plays no role.}
\resizebox{0.9\columnwidth}{!}{
\begin{tabular}{@{}lccc@{}}
\toprule
\multirow{2}{*}[-0.4em]{\textbf{Model}} & \multicolumn{3}{c}{\textbf{Rounds}} \\
\cmidrule(l){2-4} 
 & \textbf{N} & \textbf{A} & \textbf{N + A} \\
\midrule
\resnets & $2.28 \pm 0.24$ & $8.44 \pm 0.89$ & $5.66 \pm 0.68$ \\
\resnetl & $2.40 \pm 0.36$ & $8.54 \pm 0.84$ & $5.78 \pm 0.67$ \\
\vggs & $2.20 \pm 0.28$ & $8.13 \pm 0.75$ & $5.53 \pm 0.61$ \\
\bottomrule
\end{tabular}
}
\vspace{-1.75em}
\label{tab:mix-methods}
\end{table}

We first use  \sys to evaluate an MTD that combines models of the same network architecture with various training regiments. Using \sys, we first evaluate the effectiveness of the PGD attack against the naturally trained (N) and adversarially trained (A) models  only. Then, we evaluate an MTD using both the naturally trained and adversarially trained models. We set \sys to perform 100 competitions and average the results across success attacks to examine the time to fail for the defender agent. Table~\ref{tab:mix-methods} reports the average number of rounds required for the attacker to win, which can be interpreted as the robustness of the model(s) against the attack.

As we expect, the attacker always won against the naturally trained models very quickly. When switching to the adversarially trained version, the attacker required a few more steps to win, but was still able to find adversarial samples most of the time. We observe that within the same training method, the network architecture make little difference with respect to defender's robustness against the attack. Finally, the MTD that combines the naturally trained and adversarially trained models is weaker than using the adversarially trained model only. This reduced robustness is due to the presence of the naturally trained model, which was typically the first to fail in the MTD evaluations.

\subsection{MTD with Multiple Network Architectures}

\begin{table}[]
\centering
\caption{The number of rounds until the attacker succeeds in triggering misclassification by the defender using the \sys framework. The MTD consists of different combinations of naturally trained (N) and/or adversarially trained (A) \resnets, \resnetl, and \vggs models. A dash (-) indicates that the model was not used. We perform 100 trials for each experiment and report the mean number of rounds with its 95\% confidence interval. The scenarios are grouped if they are statistically equivalent. Generally, as we increase the number of adversarial models, the robustness of the MTD increases with network architecture having no effect.}
\resizebox{0.9\columnwidth}{!}{
\begin{tabular}{@{}cccc@{}}
\toprule
\multicolumn{3}{c}{\textbf{MTD Models}} & \multirow{2}{*}[-0.4em]{\textbf{Rounds}} \\
\cmidrule(lr){1-3}
\textbf{\resnets} & \textbf{\resnetl} & \textbf{\vggs} & \\
\midrule
N & N & - & $3.46 \pm 0.39$ \\
N & - & N & $3.34 \pm 0.42$ \\
- & N & N & $3.58 \pm 0.44$ \\
N & N & N & $4.64 \pm 0.70$ \\
\midrule
N & N & A & $5.68 \pm 0.69$ \\
N & A & N & $5.54 \pm 0.73$ \\
A & N & N & $5.40 \pm 0.67$ \\
\midrule
A & A & N & $6.88 \pm 0.64$ \\
A & N & A & $6.76 \pm 0.68$ \\
N & A & A & $6.84 \pm 0.61$ \\
\midrule
A & A & - & $9.34 \pm 0.85$ \\
A & - & A & $9.28 \pm 0.82$ \\
- & A & A & $9.46 \pm 0.80$ \\
A & A & A & $9.72 \pm 0.86$ \\
\bottomrule
\end{tabular}
}
\vspace{-1.75em}
\label{tab:mix-architecture}
\end{table}

We next explore MTD models using a combination of different model architectures with varying training methods. As before, we evaluate each MTD across 100 competitions and report the average number of rounds before the attacker was successful to evaluate the time to fail for the defender agent. We present the results of several combinations in Table Table~\ref{tab:mix-architecture}\footnote{Due to space restrictions, we do not include the full set of results in the main paper.}. The results have been divided into four sections based on the number of naturally or adversarially trained models used in the MTD.

Compared to the individual model results in Table~\ref{tab:mix-methods}, when combining models using the same training method, we observe a slight increase in the number of rounds required for the attacker to win. As before, using adversarially trained models only results in a higher average number of rounds for the attacker to win. We also generally observe that as the number of models in the ensemble increases and the number of adversarially trained models, the robustness of the MTD increases. However, as before, the presence of naturally trained models in the MTD reduces the average robustness of the defense due to them being most likely to fail.

\section{What causes transferability in the MTD?}

\begin{table}[]
\centering
\caption{The cosine similarity of the image perturbations between two MTD models. The \resnets and \vggs models were used for the MTD. All 2-pair combinations of naturally (N) and adversarially trained (A) models were used. We report the cosine similarity of the total perturbation from the original image averaged across all rounds of a single trial and the final image between two trials. For each experiment, a single random image was chosen from the \cifar test set such that either the attacker or defender would purposefully win. Regardless of whether the attack or defender wins, the cosine similarity remains high for both metrics.}
\resizebox{0.9\columnwidth}{!}{
\begin{tabular}{@{}ccccc@{}}
\toprule
 & \multicolumn{2}{c}{\textbf{Models}} & \multicolumn{2}{c}{\textbf{Cosine Similarity}} \\
\cmidrule(lr){2-3} \cmidrule(lr){4-5}
 & \textbf{\resnets} & \textbf{\vggs} & \textbf{Round Avg.} & \textbf{Final Image} \\
\midrule
\multirow{2}{*}[-0.8em]{Attacker}
 & N & N & .907 & .965 \\
 & N & A & .915 & .966 \\
\multirow{2}{*}[0.8em]{Wins}
 & A & N & .939 & .957 \\
 & A & A & .961 & .952 \\
\midrule
\multirow{2}{*}[-0.8em]{Defender}
 & N & N & .852 & .842 \\
 & N & A & .843 & .836 \\
\multirow{2}{*}[0.8em]{Wins}
 & A & N & .841 & .838 \\
 & A & A & .893 & .833 \\
\bottomrule
\end{tabular}
}
\vspace{-1.75em}
\label{tab:cosine-sim}
\end{table}

Based on the results in Tables \ref{tab:mix-methods} and \ref{tab:mix-architecture}, the transferability of adversarial examples described by prior work appears to a pervasive issue regardless of model architecture or training method. However, while prior work mainly focused on the transferability of adversarial examples trained on a single model, \sys shows that the transferability issue remains even when the attacker's loss gradients come from different defender models. We hypothesize that in addition to models sharing similar decision boundaries when trained on the same dataset, they also share similar loss surfaces. If so, this means that for MTD and ensemble based defenses, regardless of the models used, the attacker will always obtain a loss gradient that is similar in direction to the previous loss gradient.

To validate our hypothesis, we study the similarity of the loss gradients between the \resnets and \vggs models. We use \sys, but only execute a single competition with a maximum of 10 rounds. This round limit is set to increase the likelihood the defender wins, especially for naturally trained models. We compute an additional metric, the cosine similarity of the loss gradients. During each round, in addition to selecting a single model to respond to the attacker's query, the defender also computes the loss gradient for both the \resnets and \vggs models. It uses these loss gradients to compute the ``per round'' cosine similarity of the loss gradients. We also compute the ``final round'' similarity by randomly selecting two trials with the same final result (attacker wins or defender wins) and computing the cosine similarity between their final adversarial samples. the attacker's generated adversarial sample and computing the cosine similarity.

The results are shown in Table~\ref{tab:cosine-sim}. We observe that regardless of whether the attacker or the defender wins, the cosine similarity remains high. This observation indicates that even in cases where the attack would fail on one of the underlying models, the returned loss gradient is still useful to the attacker for finding an adversarial sample that succeeds against the other models in the MTD. This result remains consistent regardless of the training method or network architecture.

\subsection*{How can we improve MTD?}

As our experiments showed, the MTD is an ineffective method to create robust defender models. Both natural and adversarial training result in models with similar loss gradients when attacked with a PGD attacker. Our observations align with prior work which also found that ensemble-based defenses with vulnerable models were ineffective at improving adversarial robustness~\cite{he2017adversarial,tramer2020adaptive}. However, our investigation reveals a possible solution towards improving MTD. As natural and adversarial training seek to minimize the classification loss with respect to a dataset distribution, they do not consider the shaping of the loss surface as part of the training. More specifically, they do not encourage diversification of the loss gradient between models. Therefore, if, when constructing an MTD or ensemble-based defense, we use a training technique to improve gradient diversity within the defense, a PGD attacker would be less likely to succeed. As a next step, we plan on re-evaluating the MTD after training models using a gradient diversification approach~\cite{yang2021trs, lee2021graddiv}.

\section{Conclusion}

The security of AI systems with respect to adversarial evasion attacks has become an increasingly important issue due to widespread use. Sadly, a lack of proper evaluation tools discourages ML practitioners from recognizing adversarial evasion attacks as a threat and prevents researchers from properly evaluating their proposed attacks and defenses in realistic deployment scenarios. We proposed \sys, an evaluation framework designed for complex adversarial attacker-defender interactions in a more realistic environment. Unlike prior work, \sys evaluations enable dynamic interacts between the attacker and defender and allow for evaluation under multiple threat scenarios. Using \sys, we re-examined the ensemble/MTD defense architecture previously shown to be vulnerable to adversarial attacks due to the transferability property. As with prior work, the \sys evaluation demonstrated that, regardless of the underlying base models, the attacker was always successful at defeating the defender. Adversarial training, a state-of-the-art defense used in the evaluation, only hinders the attacker slightly. We performed a deeper investigation and found that the transferability of adversarial examples is mainly due to shared loss gradients between the underlying base models. These findings suggest that the previously dismissed ensemble/MTD architecture could be robust to adversarial attacks if trained to mitigate the intra-gradient similarity.

\sys is publicly available to the research community. Our development and usage of the framework as an evaluation tool is ongoing.
Although prior work demonstrated that many empirical defenses are not robust to adaptive white-box adversaries~\cite{Athalye2018ObfuscatedGG}, it remains to be seen if this fact holds true with respect to other threat scenarios. As future work, we will re-visit these ``broken'' defenses and evaluate their effectiveness with \sys under the black-box threat model. Furthermore, we plan to integrate mitigation responses external to the classification model such as detection of ongoing attack campaigns~\cite{10.1145/3385003.3410925} and observe the effect on the attacker.

\section*{Acknowledgment}

We thank the anonymous reviewers for their valuable feedback. This work was supported by the Office of Naval Research under grants N00014-20-1-2858 and N00014-22-1-2001, and Air Force Research Lab under grant FA9550-22-1-0029. Any opinions, findings, or conclusions expressed in this material are those of the authors and do not necessarily reflect the views of the sponsors.

\bibliographystyle{IEEEtran}
\bibliography{references}

 \end{document}